\pdfoutput=1

\documentclass[11pt]{article}

\usepackage{naacl2021}

\usepackage{times}
\usepackage{latexsym}
\usepackage[T1]{fontenc}
\usepackage[utf8]{inputenc}
\usepackage{microtype}
\usepackage{enumitem}
\usepackage{multirow}
\usepackage{tikz}

\usepackage{wrapfig}
\usepackage{algorithm}
\usepackage{algorithmic}

\usepackage{cancel}

\usepackage[normalem]{ulem}
\renewcommand{\algorithmiccomment}[1]{\hfill$\triangleright$\textit{#1}}

\usepackage{tikz}
\usetikzlibrary{positioning,arrows,trees,shapes,fit}

\usepackage{amsmath}
\usepackage{amsfonts,bm}

\newtheorem{prop}{Proposition}

\newcommand{\sarrow}[1][4pt]{\!\mathrel{%
   \vcenter{\hbox{\rule[-.5\fontdimen8\textfont3]{#1}{\fontdimen8\textfont3}}}%
   \mkern-4mu\hbox{\usefont{U}{lasy}{m}{n}\symbol{41}}}\!}
\makeatletter   
\newcommand{\sveryshortarrow}[1][3pt]{\mathrel{%
    \vcenter{\hbox{\rule[-.5\fontdimen8\scriptfont3]
               {\scriptratio\dimexpr#1\relax}{\fontdimen8\scriptfont3}}}%
   \mkern-4mu\hbox{\let\f@size\sf@size\usefont{U}{lasy}{m}{n}\symbol{41}}}}
\makeatother

\def\eqref#1{equation~\ref{#1}}

\def\1{\bm{1}}

\def\va{{\bm{a}}}
\def\vb{{\bm{b}}}

\def\vd{{\bm{d}}}
\def\ve{{\bm{e}}}
\def\vf{{\bm{f}}}

\def\vh{{\bm{h}}}

\def\vs{{\bm{s}}}

\def\vw{{\bm{w}}}
\def\vx{{\bm{x}}}

\def\vip{{\bm{i}\bm{p}}}

\def\m1{{\bm{1}}}

\def\mW{{\bm{W}}}

\DeclareMathAlphabet{\mathsfit}{\encodingdefault}{\sfdefault}{m}{sl}
\SetMathAlphabet{\mathsfit}{bold}{\encodingdefault}{\sfdefault}{bx}{n}
\newcommand{\tens}[1]{\bm{\mathsfit{#1}}}

\def\tW{{\tens{W}}}

\def\gS{{\mathcal{S}}}

\def\gST{{\mathcal{S}\mathcal{T}}}

\def\sC{{\mathbb{C}}}

\def\sL{{\mathbb{L}}}

\def\sS{{\mathbb{S}}}

\def\sSB{{\mathbb{S}\mathbb{B}}}

\newcommand{\Ls}{\mathcal{L}}

\newcommand{\softmax}{\mathrm{softmax}}

\DeclareMathOperator*{\argmax}{arg\,max}

\DeclareMathOperator{\real}{\rm I\!R}
\DeclareMathOperator*{\pop}{pop}
\DeclareMathOperator*{\push}{push}

\def\blackcheck{\tikz\fill[scale=0.4, color=black](0,.35) -- (.25,0) -- (1,.7) -- (.25,.15) -- cycle;}
\usepackage{xspace}

\newcommand{\ie}{{\em i.e.,}\xspace}
\newcommand{\eg}{{\em e.g.,}\xspace}

\newcommand{\Ni}{({\em i})~}
\newcommand{\Nii}{({\em ii})~}

\usepackage{url}
\usepackage{hyperref}
\usepackage{url}
\usepackage{graphicx}
\usepackage{amssymb}
\usepackage{comment}

\usepackage{mathtools}

\usepackage{caption}
\usepackage{subcaption}
\usepackage{booktabs}
\DeclareMathAlphabet{\pazocal}{OMS}{zplm}{m}{n}
\DeclareMathAlphabet{\pazocal}{OMS}{zplm}{m}{n}

\usepackage[nameinlink]{cleveref}
\crefformat{section}{\S#2#1#3} 
\crefname{algorithm}{Alg.}{Algs.}
\crefformat{subsection}{\S#2#1#3}
\Crefname{equation}{Eq.}{Eqs.}
\Crefname{figure}{Fig.}{Figs.}

\title{RST Parsing from Scratch}

\author{Thanh-Tung Nguyen$^{\dagger}$$^\P$, Xuan-Phi Nguyen$^{\dagger}$$^\P$, Shafiq Joty$^\P$$^\S$, Xiaoli Li$^\dagger$$^\P$\\
  $^\P$Nanyang Technological University \\
  $^\S$Salesforce Research Asia \\
  $^\dagger$Institute for Infocomm Research, A-STAR \\
  Singapore \\
  \texttt{\{ng0155ng@e.;nguyenxu002@e.;srjoty@\}ntu.edu.sg} 
  \\
  \texttt{xlli@i2r.a-star.edu.sg} 
}

\date{}

\begin{document}
\maketitle

\begin{abstract}
We introduce a novel top-down end-to-end formulation of document level discourse parsing in the Rhetorical Structure Theory (RST) framework. In this formulation, we consider discourse parsing as a sequence of splitting decisions at token boundaries and use a seq2seq network to model the splitting decisions. Our framework facilitates discourse parsing from scratch without requiring discourse segmentation as a prerequisite; rather, it yields segmentation as part of the parsing process. Our unified parsing model adopts a beam search to decode the best tree structure by searching through a space of high scoring trees. With extensive experiments on the standard English RST discourse treebank, we demonstrate that our parser outperforms existing methods by a good margin in both end-to-end parsing and parsing with gold segmentation. More importantly, it does so without using any handcrafted features, making it faster and easily adaptable to new languages and domains.    
\end{abstract}

\section{Introduction} \label{sec:intro}
In a document, the clauses, sentences and paragraphs are logically connected together to form a coherent discourse. The goal of discourse parsing is to uncover this underlying coherence structure, which has been shown to benefit numerous NLP applications including  text classification \cite{ji-smith-2017-neural}, summarization \cite{gerani-etal-2014-abstractive}, sentiment analysis \citep{bhatia-etal-2015-better}, machine translation evaluation \citep{joty-etal-2017-discourse} and  
conversational machine reading \citep{gao-etal-2020-discern}.

\begin{figure*}[t!]
\begin{tikzpicture}
\node[align=left, above] at (-1,0)
{\includegraphics[width=0.95\textwidth]{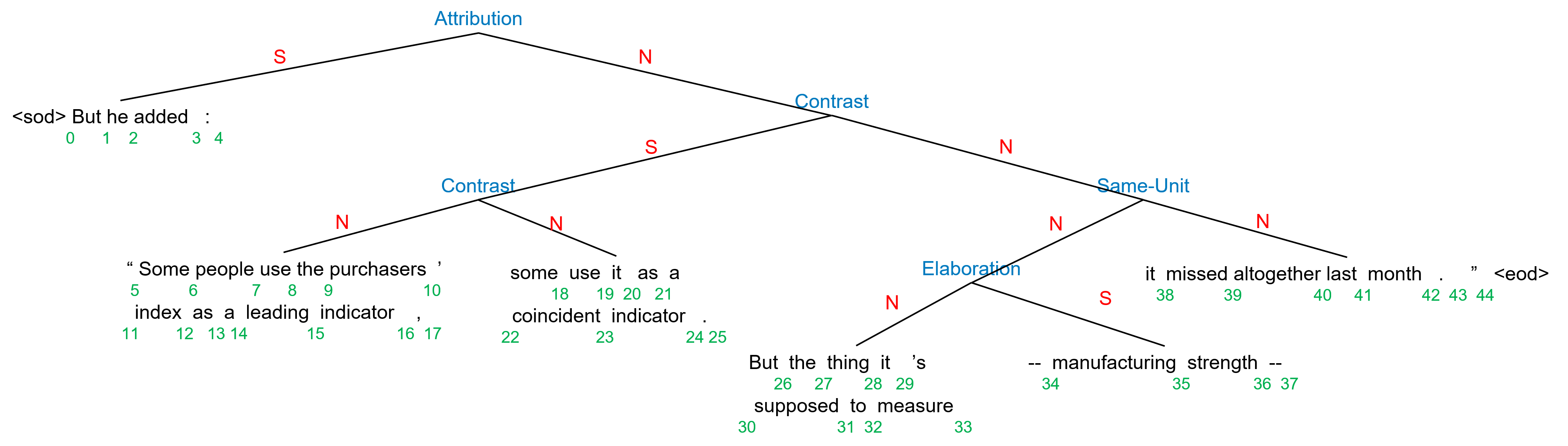}};
\small
\node[align=left, above] at (-3.7,-0.5) {\textbf{Boundary-based splitting representation when EDUs are provided}};     
\node[align =left, above] at (-2.2,-1) {\textbf{$\sS_{\text{edu}} =$} \{$(0,44) \sarrow 4$, $(4,44)\sarrow 25$, $(4,25)\sarrow 17$,$(25,44)\sarrow 37$, $(25,37)\sarrow 33$\}};
\node[align=left, above] at (-3.8,-1.5) {\textbf{Boundary-based splitting representation for end-to-end parsing}};    
\node[align =left, above] at (-1.6,-2.3) {\textbf{$\sS =$} \{$(0,44)\sarrow 4$, $\mathbf{(0,4)\sarrow 4}$, $(4,44)\sarrow 25$, $(4,25)\sarrow 17$, $\mathbf{(4,17) \sarrow 17}$, $\mathbf{(17,25)\sarrow 25}$, \\
$~~~~~~~~~~(25,44)\sarrow 37$, $(25,37)\sarrow 33$, $\mathbf{(25,33)\sarrow 33}$, $\mathbf{(33,37)\sarrow 37}$, $\mathbf{(37,44)\sarrow 44}$\}};        
\normalsize
\end{tikzpicture}
\caption{A discourse tree for two sentences in the RST discourse treebank. The internal nodes (\eg\ \emph{Attribution}, \emph{Contrast}) denote the coherence relations and the edge labels reflect the nuclearity of the child span. Below the tree, we show the sequence of splitting decisions $\sS_{\text{edu}}$ when EDUs are provided and $\sS$ when EDUs are not provided (end-to-end parsing). The \textbf{bold} splitting decision represents the final split of the span, forming an EDU.
} 
\label{fig:Discourse2SplittingFormat}
\end{figure*}

Rhetorical  Structure Theory or RST \citep{mann1988rhetorical}, one of the most influential theories of discourse, postulates a hierarchical discourse structure called discourse tree (DT). The leaves of a DT are clause-like units, known as elementary discourse units (EDUs). Adjacent EDUs and higher-order spans are connected hierarchically through coherence relations (\eg\ \emph{Contrast, Explanation}). Spans connected through a relation are categorized based on their relative importance --- \emph{nucleus} being the main part, with \emph{satellite} being the subordinate one. \Cref{fig:Discourse2SplittingFormat} exemplifies a DT spanning over two sentences and six EDUs. Finding discourse structure generally requires breaking the text into EDUs (discourse segmentation) and linking the EDUs into a DT (discourse parsing).

Discourse parsers can be singled out by whether they apply a  bottom-up or top-down procedure. Bottom-up parsers include transition-based models \citep{feng-hirst-2014-linear,ji-eisenstein-2014-representation,braud-etal-2017-cross,wang-etal-2017-two} or globally optimized chart parsing models \citep{soricut-marcu-2003-sentence,joty-carenini-ng-mehdad-acl-13,joty-etal-2015-codra}. The former constructs a DT by a sequence of shift and reduce decisions, and can parse a text in asymptotic running time that is linear in number of EDUs. However, the transition-based parsers make greedy local decisions at each decoding step, which could propagate errors into future steps. In contrast, chart parsers learn scoring functions for sub-trees and adopt a CKY-like algorithm to search for the highest scoring tree. These methods normally have higher accuracy but suffer from a slow parsing speed with a complexity of $\mathcal{O}(n^3)$ for $n$ EDUs. The top-down parsers are relatively new in discourse  \cite{lin-etal-2019-unified, zhang-etal-2020-top, Kobayashi2020TopDownRP}. These methods focus on finding splitting points in each iteration to build a DT. However, the local decisions could still affect the performance as most of the methods are still greedy.

Like most other fields in NLP, language parsing has also undergone a major paradigm shift from traditional feature-based statistical parsing to end-to-end neural parsing. Being able to parse a document end-to-end from scratch is appealing for several key reasons. First, it makes the overall development procedure easily adaptable to new languages, domains and tasks by surpassing the expensive feature engineering step that often requires more time and domain/language expertise. Second, the lack of an explicit feature extraction phase makes the training and testing (decoding) faster.  

Because of the task complexity, it is only recently that neural approaches have started to outperform traditional feature-rich methods. However, successful document level neural parsers still rely heavily on handcrafted features \citep{ji-eisenstein-2014-representation,yu-etal-2018-transition,zhang-etal-2020-top,Kobayashi2020TopDownRP}. Therefore, even though these methods adopt a neural framework, they are not ``end-to-end'' and do not enjoy the above mentioned benefits of an end-to-end neural parser.  Moreover, in existing methods (both traditional and neural), discourse segmentation is detached from parsing and treated as a prerequisite step. Therefore, the errors in segmentation affect the overall parsing performance \cite{soricut-marcu-2003-sentence,joty-carenini-ng-emnlp-12}.

In view of the limitations of existing approaches, in this work we propose an end-to-end top-down document level parsing model that:  

\begin{itemize}[leftmargin=*,itemsep=0em]
    \item Can generate a discourse tree from scratch without requiring discourse segmentation as a prerequisite step; rather, it generates the EDUs as a by-product of parsing. Crucially, this novel formulation facilitates solving the two tasks in a single neural model. Our formulation is generic and works in the same way when it is provided with the EDU segmentation.    
    
    \item Treats discourse parsing as a sequence of splitting decisions at token boundaries and uses a seq2seq pointer network \citep{VinyalsNIPS2015} to model the splitting decisions at each decoding step. Importantly, our seq2seq parsing model can adopt beam search to widen the search space for the highest scoring tree, which to our knowledge is also novel for the parsing problem.  
    
    \item Does not rely on any handcrafted features, which makes it faster to train or test, and easily adaptable to other domains and languages. 

    \item Achieves the state of the art (SoTA) with an $F_1$ score of 46.6 in the Full (label+structure) metric for end-to-end parsing on the English RST Discourse Treebank, which  outperforms many parsers that use gold EDU segmentation. With gold segmentation, our model achieves a SoTA $F_1$ score of 50.2 (Full), outperforming the best existing system by 2.1 absolute points. More imporantly, it does so without using any handcrafted features (not even part-of-speech tags). 
\end{itemize}

We make our code available at \href{https://ntunlpsg.github.io/project/rst-parser}{https://ntunlpsg.github.io/project/rst-parser}

\section{Model} \label{sec:model}

Assuming that a document has already been segmented into EDUs, following the traditional approach, the corresponding discourse tree (DT) can be represented as a set of {labeled constituents}.
\begin{equation}
\sC := \{ ((i_t, k_t, j_t), r_t)|i_t \leq k_t < j_t \}_{t=1}^{m}
\label{eq:dicourse_edu_rep}
\end{equation}
where $m = |\sC|$ is the number of internal nodes in the tree and $r_t$ is the relation label between the discourse unit containing EDUs $i_t$ through $k_t$ and the one containing EDUs $k_t+1$ through $j_t$. 

Traditionally, in RST parsing, {discourse segmentation} is performed first to obtain the sequence of EDUs, which is followed by the {parsing} process to assemble the EDUs into a labeled tree. In other words, traditionally discourse segmentation and parsing have been considered as two distinct tasks that are solved by two different models. 

On the contrary, in this work we take a radically different approach that directly starts with parsing the (unsegmented) document in a top-down manner and treats discourse segmentation as a special case of parsing that we get as a by-product. Importantly, this novel formulation of the problem allows us to solve the two problems in a single neural model. Our parsing model is generic and also works in the same way when it is fed with an EDU-segmented text. Before presenting the model architecture, we first formulate the problem as a splitting decision problem at the token level.  

\subsection{Parsing as a Splitting Decision Problem}

\begin{figure}[t!]
\begin{center}
\includegraphics[width=0.45\textwidth]{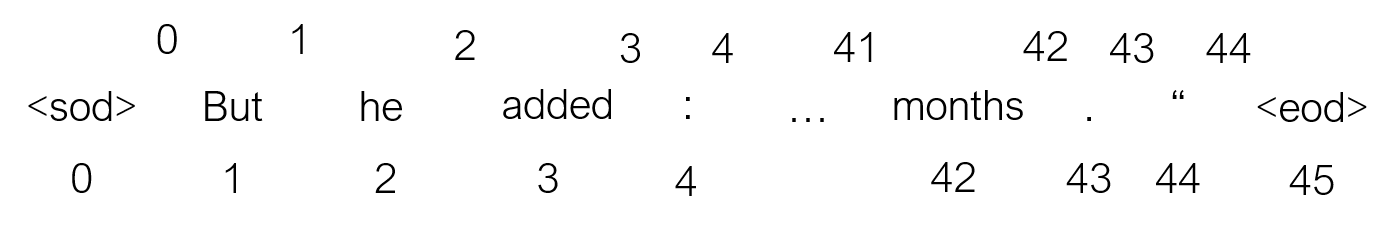}
\end{center}
\caption{Relation between token-boundary (above) and token (below) representations. A token-boundary position $k$ is located between the tokens at $k$ and $k+1$.} 
\label{fig:Token-Boundary_Convert_Docs}
\end{figure}

We reformulate the discourse parsing problem from \Cref{eq:dicourse_edu_rep} as a sequence of splitting decisions at \emph{token boundaries} (instead of EDUs). Specifically, the input text is first prepended and appended with the special start (\texttt{$<${sod}$>$}) and end (\texttt{$<${eod}$>$}) tokens, respectively. We define the token-boundary as the indexed position between two consecutive tokens. For example, the constituent spanning ``But he added :'' in \Cref{fig:Token-Boundary_Convert_Docs} is defined as $(0,4)$.

Following the standard practice, we convert the discourse tree by transforming each multi-nuclear constituent into a hierarchical right-branching binary sub-tree. Every internal node in the resulting binary tree will have a left and a right constituent, allowing us to represent it by its split into the left and right children. Based on this, we define the parsing as a set of splitting decisions $\sS$ at token-boundaries by the following proposition:
\begin{prop}
\label{prop:2} 
Given a binarized discourse tree for a document containing $n$ tokens, the tree can be converted into a set of token-boundary splitting decisions $\sS = \{(i,j)\sarrow  k| i < k \leq j\}$ such that the parent constituent $(i,j)$ either gets split into two child constituents $(i,k)$ and $(k,j)$ for $k<j$, or forms a terminal EDU unit for $k=j$, \ie\ the span will not be split further (\ie\ marks segmentation).
\end{prop}

Notice that $\sS$ is a generalized formulation of RST parsing, which also includes the decoding of EDUs as a special case ($k=j$). It is quite straight-forward to change this formulation to the parsing scenario, where discourse segmentation (sequence of EDUs) is provided. Formally, in that case, the tree can be converted into a set of splitting decisions $\sS_{\text{edu}} = \{(i,j)\sarrow  k| i < k < j\}$ such that the constituent $(i,j)$ gets split into two constituents $(i,k)$ and $(k,j)$ for $k<j$, \ie\ we simply omit the special case of $k=j$ as the EDUs are given. In other words, in our generalized formulation, discourse segmentation is just one extra step of parsing, and can be done top-down end-to-end.



An example of our formalism of the parsing problem is shown in \Cref{fig:Discourse2SplittingFormat} for a discourse tree spanning over two sentences {(44 tokens)}; for simplicity, we do not show the relation labels corresponding to the splitting decisions (marked by~$\sarrow$). Since each splitting decision corresponds to one and only one internal node in the tree, it guarantees that the transformation from the tree to $\sS$ (and $\sS_{\text{edu}}$) has a one-to-one mapping. Therefore, predicting the sequence of such splitting decisions is equivalent to predicting the discourse tree (DT).

\begin{figure*}[t!]
\centering
\includegraphics[width=0.8\textwidth]{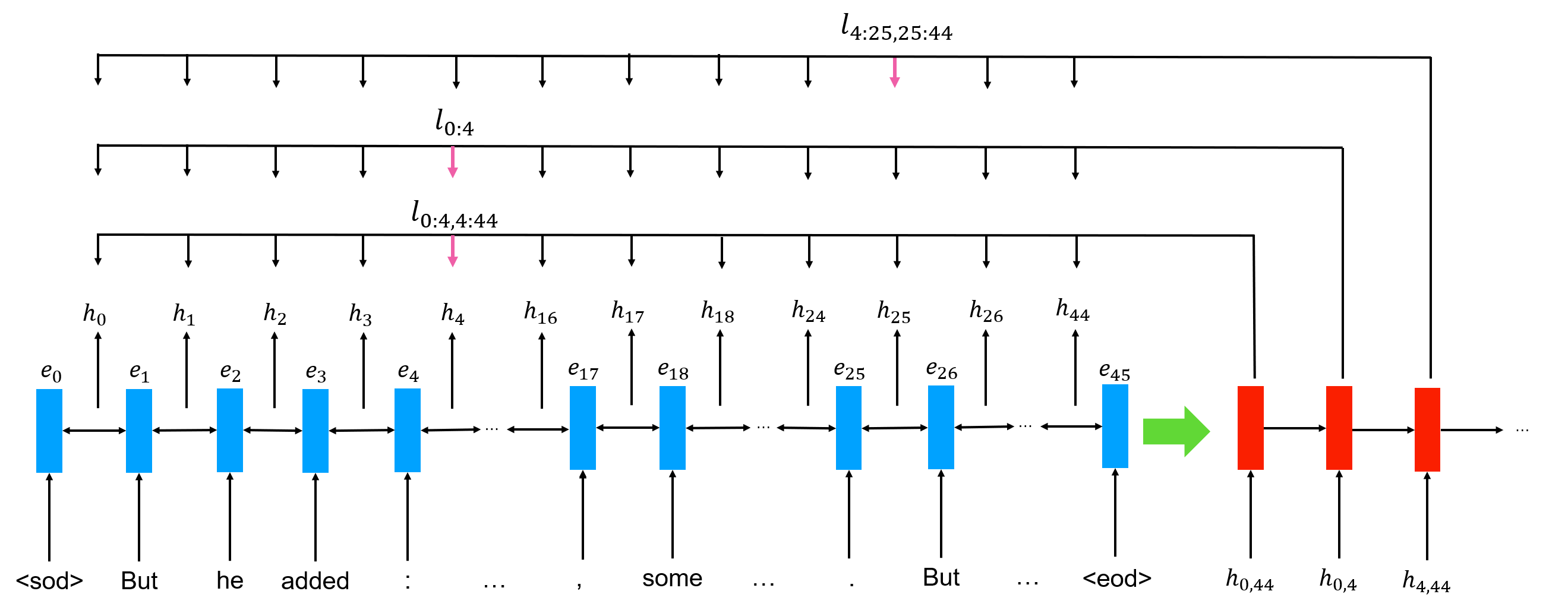}
\caption{\small Our discourse parser along with a few decoding steps for a given document. The input to the decoder at each step is the representation of the span to be split. We predict the splitting point using the biaffine function between the corresponding decoder state and the token-boundary encoder representations. The figure is for end-to-end parsing, where each EDU-corresponding span points to its right edge to mark the EDU. The coherence relations between the left and right spans are assigned using a label classifier after the (approximately) optimal tree structure is formed using beam search.}
\label{fig:discourse_parsing_base_architecture}
\end{figure*}

\paragraph{Seq2Seq Parsing Model.}

In this work, we adopt a structure-then-label framework. Specifically, we factorize the probability of a DT into the probability of the tree structure and the probability of the relations (\ie\ the node labels) as follows:

\small 
\begin{eqnarray}
 P_{\theta}(DT|\vx) =  P_{\theta}(\sS,\sL|\vx) =
   P_{\theta}(\sL|\sS,\vx) P_{\theta}(\sS|\vx) 
 \label{eq:prob_structure_label}
\end{eqnarray}
\normalsize

\noindent where $\vx$ is the input document, and $\sS$ and $\sL$ respectively denote the structure and labels of the DT. This formulation allows us to first infer the best tree structure (\eg\ using beam search), and then find the corresponding labels. 

As discussed, we consider the structure prediction problem as a sequence  of splitting decisions to generate the tree in a top-down manner. We use a seq2seq pointer network \citep{VinyalsNIPS2015} to model the sequence of splitting decisions (\Cref{fig:discourse_parsing_base_architecture}). We adopt a depth-first order of the decision sequence, which showed more consistent performance in our preliminary experiments than other alternatives, such as breath-first order.

First, we encode the tokens in a document $\vx = (x_0, \ldots, x_n$) with a document encoder and get the token-boundary representations ($\vh_0, \ldots, \vh_n$). Then, at each decoding step $t$, the model takes as input an internal node $(i_t,j_t)$, and produces an output $y_t$ (by pointing to the token boundaries) that represents the splitting decision $(i_t,j_t) \sarrow k_t$ to split it into two child constituents $(i_t,k_t)$ and $(k_t,j_t)$. For example, the initial span $(0,44)$ in  \Cref{fig:Discourse2SplittingFormat}  is split at boundary position $4$, yielding two child spans $(0,4)$ and $(4,44)$. If the span $(0,4)$ is given as an EDU (\ie\  segmentation given), the splitting stops at $(0,4)$, thus omitted in $\sS_{\text{edu}}$ (\Cref{fig:Discourse2SplittingFormat}). Otherwise, an extra decision $(0,4)\sarrow 4 \in \sS$ needs to be made to mark the EDUs for end-to-end parsing. With this,
the probability of $\sS$ can be expressed as:

\small 
\begin{eqnarray}
\hspace{-0.3em} P_{\theta}(\sS|\vx) \hspace{-0.5em} &=& \hspace{-0.5em}\prod\limits_{y_t \in \sS} P_{\theta}(y_t | y_{<t},\vx) \nonumber \\
\hspace{-2em} &=& \hspace{-0.5em} \prod \limits_{t=1}^{|\sS|} P_{\theta} \Big( (i_t,j_t)\sarrow k_t | ((i,j)\sarrow k)_{<t},\vx \Big) 
\nonumber
\label{eq:prob_structure}
\end{eqnarray}
\normalsize

\noindent This end-to-end conditional splitting formulation is the main novelty of our method and is in  contrast to previous approaches which rely on offline-inferred EDUs from a separate discourse segmenter. Our formalism streamlines the overall parsing process, unifies the neural components seamlessly and smoothens the training process. 

\subsection{Model Architecture} 

In the following, we describe the components of our parsing model: the document encoder, the boundary and span representations, the decoding process through the decoder and the label classifier.

\paragraph{Document Encoder.}
Given an input document of $n$ words $\vx = (x_1, \ldots, x_n)$, we first add \texttt{$<${sod}$>$} and \texttt{$<${eod}$>$} markers to the  sequence. After that, each token $x_i$ in the sequence is mapped into its dense vector representation $\ve_i$ as: $\ve_i = [\ve_i^{\text{char}} , \ve_i^{\text{word}} ]$, where $\ve_i^\text{char}$, and $\ve_i^\text{word}$ are respectively the character and word embeddings of token $x_i$.  For word embedding, we experiment with \Ni randomly initialized, \Nii pretrained static embeddings ,\eg\ GloVe \citep{pennington2014glove}). To represent the character embedding of a token, we apply a character bidirectional LSTM \ie\ Bi-LSTM \citep{Hochreiter:1997} or pretrained contextualized embeddings, \eg\ XLNet \citep{NEURIPS2019_dc6a7e65}.  The token representations are then passed to a sequence encoder of a three-layer Bi-LSTM  to obtain their forward $\vf_i$ and backward $\vb_i$ contextual representations. 

\paragraph{Token-boundary Span Representations.}

To represent each token-boundary position $k$ between token positions $k$ and $k+1$, we use the fencepost representation \citep{cross-huang-2016-span}:
\begin{equation}
\small
    \vh_{k} = [\vf_{k};\vb_{k+1}]
\end{equation}
where $\vf_k$ and $\vb_{k+1}$ are the forward and backward LSTM hidden vectors of positions $k$ and $k+1$ respectively, and $[\cdot;\cdot]$ is the concatenation operation. 

\begin{figure}[t!]
\centering
\includegraphics[width=0.4\textwidth]{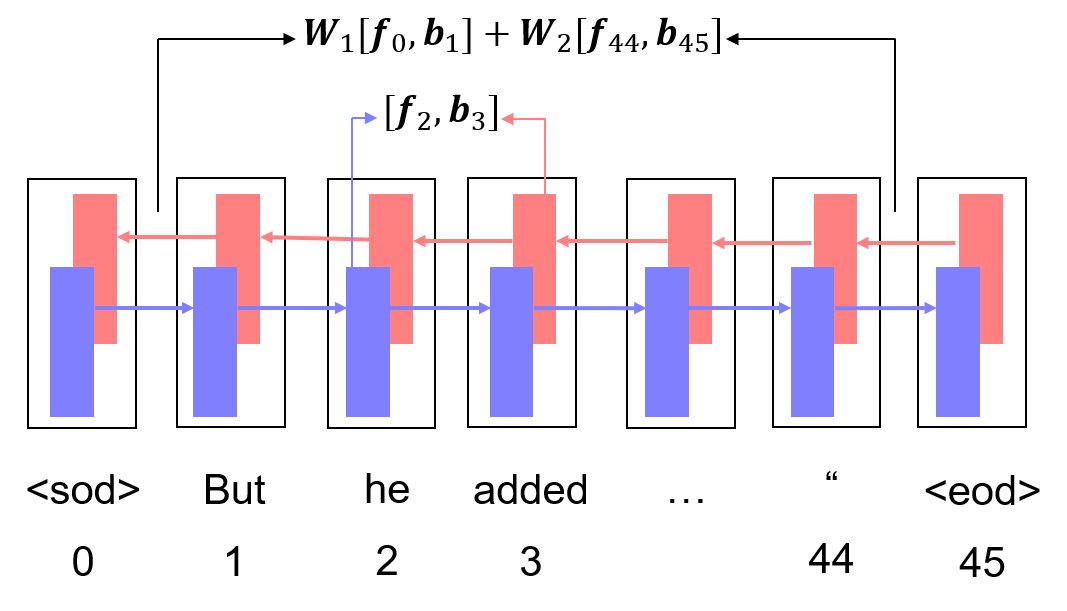}
\caption{\small Illustration of token-boundary span encoder. The figure lays out an example representation for the boundary at $2$ and the representation of the token-boundary span $(0, 44)$, which corresponds to the whole document.}
\label{fig:boundary_span_representation}
\end{figure}

Then, to represent the token-boundary span $(i,j)$, we use the linear combination of the two endpoints $i$ and $j$ as: 
\begin{equation}
\small 
    \vh_{i,j} = \mW_1 \vh_i + \mW_2 \vh_j
\end{equation}
where $\mW_1$ and $\mW_2$ are trainable weights. These span representations will be used as input to the decoder or the label classifier. \Cref{fig:boundary_span_representation} illustrates an example boundary span representation.

\paragraph{The Decoder.}

Our model uses a unidirectional LSTM as the decoder. At each decoding step $t$, the decoder takes as input the corresponding span $(i,j)$ (\ie\ $\vh_{i,j}$) and its previous LSTM state $\vd_{t-1}$ to generate the current state $\vd_t$ and then the biaffine function \citep{DozatM17} is applied between $\vd_t$ and \emph{all} the encoded token-boundary representations $(\vh_0, \vh_1, \ldots, \vh_n)$ as follows:

\small 
\begin{eqnarray}
\vd'_t =  \text{MLP}_d(\vd_t) \hspace{1em} \vh'_i = \text{MLP}_h(\vh_i) \\
s_{t}^{i} = {\vd'_t}^T \mW_{dh} \vh'_i + {\vh'_i}^T\vw_h\\
a_t^i = \frac{\exp(s_{t}^{i})}{\sum_{i=0}^n \exp(s_{t}^{i})} \hspace{1em} \text{ for } i= 0, \ldots, n
\label{eq:decoder_pointing}
\end{eqnarray}
\normalsize

\noindent where each $\text{MLP}$ operation comprises a linear transformation with LeakyReLU activation \citep{Maas13rectifiernonlinearities} to transform $\vd_i$ and $\vh_i$ into equal-sized vectors $\vd'_t, \vh'_i \in \real^d$, and $\mW_{dh} \in \real^{d  \times d}$ and $\vw_h \in \real^d$ are respectively the weight matrix and weight vector for the biaffine function. The resulting biaffine scores $s_{t}^i$ are then fed into a \emph{softmax} layer to acquire the pointing distribution $\va_t^i \in [0,1]^{n+1}$ for the splitting decision.
During inference, when decoding the tree at step $t$, we only examine the ``valid'' splitting points between $i$ and $j$, and we look for $k$ such that $i<k \leq j$.

\paragraph{Label Classifier.}

We perform label assignment after decoding the entire tree structure. Each assignment takes into account the splitting decision that generated it since the label represents the relation between the child spans.
Specifically, for a constituent $(i,j)$ that was split into two child constituents $(i,k)$ and $(k,j)$, we determine the coherence relation between them as follows:

\small 
\begin{eqnarray}
\vh^l_{ik} =  \text{MLP}_l([\vh_i;\vh_k]); ~~ \vh^r_{kj} = \text{MLP}_r([\vh_k;\vh_j]) \\
P_{\theta}(l|(i,k),(k,j)) = \softmax \Big((\vh^l_{ik})^T \tW_{lr} \vh^r_{kj} \nonumber \\
+ (\vh^l_{ik})^T\mW_l + (\vh^r_{kj})^T\mW_r + \vb \Big) \\
l^*_{(i,k),(k,j)}= \argmax_{l \in L}P_{\theta}(l|(i,k),(k,j))
\label{eq:discourse_labelling}
\end{eqnarray}
\normalsize

\noindent where $L$ is the total number of labels (\ie\ coherence relations with nuclearity attached); each of $\text{MLP}_l$ and $\text{MLP}_r$ includes a linear transformation with LeakyReLU activation to transform the left and right spans into equal-sized vectors $\vh^l_{ik}, \vh^r_{kj} \in \real^d$; $\tW_{lr}\in \real^{d \times L \times d}, \mW_l \in \real^{d \times L},\mW_r\in \real^{d \times L}$ are the weights and $\vb$ is a bias vector. 


\paragraph{Training Objective.}
Our parsing model is trained by minimizing the total loss defined as:
\begin{equation}
\small 
\Ls (\theta_e,\theta_d,\theta_l) = \Ls_{s} (\theta_e,\theta_d) + \Ls_{l} (\theta_e, \theta_{l})
\end{equation}
where structure $\Ls_{s}$ and label $\Ls_{l}$ losses are cross-entropy losses computed for the splitting and labeling tasks respectively, and $\theta_e$, $\theta_{d}$ and $\theta_{l}$ denote the encoder, decoder and labeling parameters. 

\subsection{Complete Discourse Parsing Models}\label{sec:complete_models}
Having presented the generic framework, we now describe how it can be easily adapted to the two parsing scenarios: \Ni end-to-end parsing and \Nii parsing with EDUs. We also describe the incorporation of beam search for inference.

\paragraph{End-to-End Parsing.} As mentioned, previous work for end-to-end parsing assumes a separate segmenter that provides EDU-segmented texts to the parser. Our method, however, is an end-to-end framework that produces both the EDUs as well as the parse tree in the same inference process. To guide the search better, we incorporate an inductive bias into our inference based on the finding that most sentences have a well-formed subtree in the document-level tree \citep{soricut-marcu-2003-sentence}, \ie\ discourse structure tends to align with the text structure (sentence boundary in this case); for example,  \citet{fisher-roark-2007-utility,joty-carenini-ng-mehdad-acl-13} found that more than $95\%$ of the sentences have a well-formed subtree in the RST discourse treebank.

Our goal is to ensure that each sentence corresponds to an internal node in the tree. This can be achieved by a simple adjustment in our inference. When decoding at time step $t$ with the span $(i_t, j_t)$ as input, if the span contains $M >0$ sentence boundaries within it, we pick the one that has the highest pointing score (Eq. \ref{eq:decoder_pointing}) among the $M$ alternatives as the split point $k_t$. If there is no sentence boundary within the input span ($M=0$), we find the next split point as usual. In other words, sentence boundaries in a document get the chance to be split before the token boundaries inside a sentence. This constraint is indeed similar to the 1S-1S (1 subtree for 1 sentence) constraint of \citet{joty-carenini-ng-mehdad-acl-13}'s bottom-up parsing, and is also consistent with the property that EDUs are always within the sentence boundary. Algorithm \ref{alg:end-to-end-parsing} illustrate the end-to-end inference algorithm.

\begin{algorithm}[t!]
\scriptsize
    \captionsetup{font=scriptsize}
    \caption{Discourse Tree Inference (end-to-end)}
    \label{alg2}
    \begin{algorithmic}
    \REQUIRE Document length $n$; boundary encoder states: $(\vh_0, \vh_1, \ldots, \vh_n)$; sentence boundary set $\sSB$ ;  label scores: $P(l|(i,k), (k,j))$, $0\leq i < k \leq j \leq n, l \in L$, initial decoder state $st$.
    \ENSURE Parse tree ${DT}$
    \STATE $\gST=[(0,n)]$  \algorithmiccomment{stack of spans}
    \STATE $\gS=[]$
    \WHILE {$\gST \neq \varnothing$}
        \STATE $(i, j) = \pop(\gST)$ \algorithmiccomment{Current span to split}
        \STATE $\va, st=dec(st, \vh_{i,j})$ \algorithmiccomment{$\va$: split prob. dist.}
         \IF{$(i,j) \cap \sSB \ne \varnothing$} 
            \STATE $k=\argmax_{i<k \leq j\ \&\ k \in \sSB}\va $
        \ELSE
            \STATE $k=\argmax_{i<k \leq j}\va $
        \ENDIF
        
                \IF{$j-1 > k > i+1$}    
                    \STATE $\push(\gST,(k,j))$
                    \STATE $\push(\gST,(i,k))$
                \ELSIF{$j-1 > k = i+1 $}    
                    \STATE $\push(\gST,(k,j))$
                \ELSIF{$k=j-1 > i+1$}    
                    \STATE $\push(\gST,(i,k))$
                \ENDIF
                \IF{$k \neq j$}    
                    \STATE $\push(\gS((i,k,j))$
                \ENDIF
    \ENDWHILE
    \STATE ${DT}=[((i,k,j),argmax_{l}P(l|(i,k)(k,j)) \forall (i,k,j) \in \gS]$
  \end{algorithmic}
\label{alg:end-to-end-parsing}
\end{algorithm}

\paragraph{Parsing with EDUs.} When segmentation information is provided, we can have a better encoding of the EDUs to construct the tree. Specifically, rather than simply taking the token-boundary representation corresponding to the EDU boundary as the EDU representation, we adopt a hierarchical approach, where we add another Bi-LSTM layer (called ``Boundary LSTM'') that connects EDU boundaries (a figure of this framework is in the Appendix). In other words, the input sequence to this LSTM layer is $(\overline{\vh}_0,\ldots,\overline{\vh}_m)$, where $\overline{\vh}_0=\vh_0$, $\overline{\vh}_m=\vh_n$ and $\overline{\vh}_j \in \{\vh_1,\ldots,\vh_{n-1}\}$ such that $\overline{\vh}_j$ is an EDU boundary. For instance, for the example in \Cref{fig:Discourse2SplittingFormat}, the input to the Boundary LSTM layer is $(\vh_0,\vh_4,\vh_{17},\vh_{25},\vh_{33},\vh_{37},\vh_{44})$. 

This hierarchical representation facilitates better modeling of relations between EDUs and higher order spans, and can capture long-range dependencies better, especially for long documents.

\paragraph{Incorporating Beam Search.} 

Previous work \citep{lin-etal-2019-unified, zhang-etal-2020-top} which also uses a seq2seq architecture, computes the pointing scores over the token or span representations only within the input span. For example, for an input span $(i,j)$, the pointing scores are computed considering only $(\vh_i, \ldots, \vh_j)$ as opposed to $(\vh_1, \ldots, \vh_n)$ in our Eq. \ref{eq:decoder_pointing}. This makes the scales of the scores uneven across different input spans as the lengths of the spans vary. Thus, such scores cannot be objectively compared across sub-trees globally at the full-tree level. In addition, since efficient global search methods like beam search cannot be applied properly with non-uniform scores, these previous methods had to remain greedy at each decoding step. In contrast, our decoder points to all the encoded token-boundary representations in every step (Eq. \ref{eq:decoder_pointing}). This ensures that the pointing scores are evenly scaled, allowing fair comparisons between the scores of all candidate sub-trees. Therefore, our method enables the effective use of beam search through highly probable candidate trees. Algorithm \ref{alg:parsing-with-EDUs} illustrates the beam search inference when EDUs are given.

\begin{algorithm}[t!]
\scriptsize
    \captionsetup{font=scriptsize}
    \caption{Discourse Tree Inference with Beam Search (with gold EDUs)}
    \label{alg1}
    \begin{algorithmic}
    \REQUIRE Number of EDUs in document $m$; beam width $B$; EDU boundary-based encoder states: $(\overline{\vh}_0,\ldots,\overline{\vh}_m)$; label scores: $P_{\theta} (l|(i,k),(k,j)$, $0\leq i < k < j \leq m, l \in \{1, \ldots, L\}$, initial decoder state $\vs$.
    \ENSURE Parse tree ${DT}$
    \STATE $L_d = m-1$ \algorithmiccomment{Decoding length}
    \STATE {beam} = array of $L_d$ items            \algorithmiccomment{List of empty beam items}
    \STATE init\_input\_span$=[(0,m),(0,0),\ldots, (0,0)]$ \algorithmiccomment{$m$-$2$ paddings (0,0)}
    \STATE init\_tree$=[(0,0,0),(0,0,0),\ldots, (0,0,0)]$ \algorithmiccomment{$m$-$1$ elements}
    \STATE beam{[0]} = $(0, \vs, init\_input\_span, \text{init\_tree})$ \algorithmiccomment{Initialize first item (log-prob,state,input\_span,tree)}
    \FOR {$t=1\ \TO\ L_d$}
        \FOR {$(\text{logp}, \vs, \text{input-span},\text{tree}) \in \text{beam}[t-1]$} 
            \STATE $(i,j) =\text{input-span}[t-1]$ \algorithmiccomment{Current span to split}
            \STATE $\va, \vs'= \text{decoder-step}(\vs, \overline{\vh}_{i,j})$ \algorithmiccomment{$\va$: split prob. dist.}
            \FOR {$(k, p_k) \in \text{top-}B(\va)\ \AND\ i<k<j $}
                \STATE $\text{curr-input-span}=\text{input-span}$
                \STATE $\text{curr-tree}=\text{tree}$
                \STATE $\text{curr-tree}[t-1]=(i,k,j)$
                \IF{$k > i+1$}    
                    \STATE $\text{curr-input-span}[t]=(i,k)$
                \ENDIF
                \IF{$j > k+1$}    
                    \STATE $\text{curr-input-span}[t+j-k-1]=(k,j)$
                \ENDIF
                \STATE push ({logp} + $\log(p_k), \vs',$ {curr-input-span}, {curr-tree}) to beam[t]
            \ENDFOR
        \ENDFOR
        \STATE prune beam[t] \algorithmiccomment{Keep top-$B$ highest score trees}
    \ENDFOR
    \STATE $\text{logp*},\vs^*,\vip^*,\sS^* = \argmax_{\text{logp}} \text{beam}[L_d]$ \algorithmiccomment{$\sS^*$: best structure}
    \STATE $\text{DT} =[(i,k,j, \argmax_{l} P_{\theta} (l|(i,k),(k,j)) ~~\forall (i,k,j) \in \gS^*]$
  \end{algorithmic}
\label{alg:parsing-with-EDUs}
\end{algorithm}

\section{Experiments}\label{sec:experiment}
We conduct our experiments on discourse parsing with and without gold segmentation. We use the standard English RST Discourse Treebank or RST-DT \citep{RST:DT:2002} for training and evaluation. It consists of 385 annotated Wall Street Journal news articles: 347 for training and 38 for testing. We randomly select 10\% of the training set as our development set for hyper-parameter tuning. Following prior work, we adopted the same 18  courser relations defined in \citep{carlson-marcu-01}. For evaluation, we report the standard metrics Span, Nuclearity, Relation and Full F1 scores, computed using the standard Parseval \citep{morey-etal-2017-much,morey-etal-2018-dependency} and RST-Parseval \cite{marcu-2000-rhetorical} metrics.

\subsection{Parsing with Gold Segmentation}\label{sec:parse_with_gold}

\paragraph{Settings.}

Discourse parsing with gold EDUs has been the standard practice in many previous studies. We compare our model with ten different baselines as shown in Table \ref{table:gold_segmentation_results}. We report most results from \citet{morey-etal-2018-dependency,zhang-etal-2020-top,Kobayashi2020TopDownRP}, while we reproduce \citet{yu-etal-2018-transition} using their provided source code.

For our model setup, we use the encoder-decoder framework with a 3-layer Bi-LSTM encoder and 3-layer unidirectional LSTM decoder. The LSTM hidden size is 400, the word embedding size is 100 for random initialization, while the character embedding size is 50. The hidden dimension in MLP modules and biaffine function for structure prediction is 500. The beam width $B$ is 20. Our model is trained by Adam optimizer \citep{KingmaB14} with a batch size of 10000 tokens. Our learning rate is initialized at $0.002$ and scheduled to decay at an exponential rate of $0.75$ for every 5000 steps. Model selection for testing is performed based on the Full F1 score on the development set. When using pretrained word embeddings, we use the 100D vectors from GloVe \citep{pennington2014glove}. For pretrained model, we use the \emph{XLNet-base-cased} version  \citep{NEURIPS2019_dc6a7e65}.\footnote{Our initial attempt with BERT did not offer significant gain as BERT is not explicitly designed to process long documents and has a limit of maximum $512$ tokens.} The pretrained models/embeddings are kept frozen during training.

\begin{table}[t!]
\begin{small}
\begin{center}
\begin{tabular}{l | p{0.4cm} p{0.4cm} p{0.4cm} p{0.5cm}}
\toprule
\textbf{Systems} & Span & Nuc & Rel & Full\\
\midrule
\multicolumn{5}{c}{\textbf{Parseval Metric} \citep{morey-etal-2017-much}} \\[+0.3em]
\textbf{Human Agreement} & 78.7 & 66.8 & 57.1 & 55.0  \\
\hline 
\citet{ji-eisenstein-2014-representation}$^{+}$  & 64.1  & 54.2  & 46.8 & 46.3 \\
\citet{feng-hirst-2014-linear}$^{+}$  &68.6  &55.9  &45.8  &44.6 \\
\citet{joty-etal-2015-codra}$^{+}$ &65.1  &55.5  &45.1  &44.3 \\
\citet{li-etal-2016-discourse}$^{+}$  &64.5 &54.0 &38.1 &36.6  \\
\citet{braud-etal-2016-multi}  &59.5 &47.2 &34.7 &34.3  \\
\citet{braud-etal-2017-cross}$^{*}$  &62.7 &54.5 &45.5 &45.1  \\
\citet{yu-etal-2018-transition}$^{+\S}$  &71.4 &60.3 &49.2 &48.1  \\
\citet{zhang-etal-2020-top}$^{+}$  & 67.2  &55.5  &45.3  &44.3 \\

Our with GloVe  &71.1  & 59.6   &47.7   & 46.8\\
Our with XLNet$^{\S}$  &\textbf{74.3}  & \textbf{64.3}   & \textbf{51.6}   & \textbf{50.2} \\
\midrule
\multicolumn{5}{c}{\textbf{RST-Parseval Metric} \citep{marcu-2000-rhetorical}} \\[+0.3em]
\textbf{Human Agreement} & 88.7 &77.3 &65.4\\
\hline 
\citet{yu-etal-2018-transition}$^{+\S}$  &85.5 &73.1 &60.2  \\
\citet{wang-etal-2017-two}$^{+\S}$& 86.0 & 72.4 & 59.7\\
\citet{Kobayashi2020TopDownRP}$^{+\S}$ & 87.0  & 74.6  & 60.0 \\
Our with XLNet$^{\S}$  &\textbf{87.6}  & \textbf{76.0}   & \textbf{61.8}   \\
\bottomrule
\end{tabular}
\end{center}
\caption{\small Parsing results with gold segmentation. The sign $^{+}$ denotes that systems use handcrafted features such as lexical, syntactic, sentence/paragraph boundary features and so on, $^{*}$ denotes that systems use external cross-lingual features and $^{\S}$ means that systems use pretrained models.}
\label{table:gold_segmentation_results}
\end{small}
\end{table}

\paragraph{Results.}

From the results in \Cref{table:gold_segmentation_results}, we see that our model with GloVe (static) embeddings achieves a Full F1 score of 46.8, the highest among all the parsers that do not use pretrained models (or contextual embeddings). This suggests that a BiLSTM-based parser can be competitive with effective modeling. The model also outperforms the one proposed by \citet{zhang-etal-2020-top}, which is closest to ours in terms of modelling, by 3.9\%, 4.1\%, 2.4\% and 2.5\% absolute in Span, Nuclearity, Relation and Full, respectively. More importantly, our system achieves such results without relying on external data or features, in contrast to previous approaches. In addition, by using XLNet-base pretrained model, our system surpasses all existing methods (with or without pretraining) in all four metrics, achieving the state of the art with 2.9\%, 4.0\%, 2.4\% and 2.1\% absolute improvements. It also reduces the gap between system performance and human agreement. {When evaluated with the RST-Parseval \citep{marcu-2000-rhetorical} metric, our model outperforms the baselines by 0.6\%, 1.4\% and 1.8\% in Span, Nuclearity and Relation, respectively.}

\subsection{End-to-end Parsing} \label{sec:parse_without_gold}
For end-to-end parsing, we compare our method with the model proposed by \citet{zhang-etal-2020-top}. Their parsing model uses the EDU segmentation from \citet{ijcai2018-579}. Our method, in contrast, predicts the EDUs along with the discourse tree in a unified process (\Cref{sec:complete_models}). In terms of model setup, we use a setup identical to the experiments with gold segmentation (\Cref{sec:parse_with_gold}).

\begin{table}[t!]
\begin{small}
\begin{center}
\begin{tabular}{l p{0.4cm} p{0.4cm} p{0.4cm} p{0.5cm}}
\toprule
Model  & Span & Nuc & Rel & Full \\
\midrule
\citet{zhang-etal-2020-top}  & 62.3  &50.1   &40.7   &39.6\\
\midrule
\textbf{Our model} \\
\hspace{1em} with GloVe  & 63.8  &53.0   &43.1   &42.1\\
\hspace{1em} with XLNet  & \textbf{68.4}  &\textbf{59.1}   &\textbf{47.8}   &\textbf{46.6}\\
\bottomrule
\end{tabular}
\end{center}
\caption{End-to-end parsing performance.}
\label{tb_end2end_performance}
\end{small}
\end{table}

\Cref{tb_end2end_performance} reports the performance for document-level end-to-end parsing. Compared to \citet{zhang-etal-2020-top}, our model with GloVe embeddings yields 1.5\%, 2.9\%, 2.4\% and 2.5\% absolute gains in Span, Nuclearity, Relation and Full F1 scores, respectively. Furthermore, the model with XLNet achieves even better performance and outperforms many models that use gold segmentation (Table \ref{table:gold_segmentation_results}). 

\paragraph{EDU Segmentation Results.} 

Our end-to-end parsing method gets an F1 score of 96.30 for the resulting EDUs. Our result rivals existing SoTA segmentation methods -- { 92.20 F1 of \newcite{ijcai2018-579} and 95.55 F1 of \newcite{lin-etal-2019-unified}.} This shows the efficacy of our unified framework for not only discourse parsing but also segmentation.\footnote{We could not compare our segmentation results with the {DISRPT} 2019 Shared Task \cite{zeldes-etal-2019-disrpt} participants. We found few inconsistencies in the settings. First, in their ``gold sentence'' dataset, instead of using the gold sentence, they pre-process the text with an automatic tokenizer and sentence segmenter. Second, in the evaluation, under the same settings, they do not exclude the trivial BeginSegment label at the beginning of each sentence which we exclude in evaluating our segmentation result (following the RST standard).}

\subsection{Ablation Study}
To further understand the contributions from the different components of our unified parsing framework, we perform an ablation study by removing selected components from a network trained with the best set of parameters.

\paragraph{With Gold Segmentation.}

\Cref{table:gold_edu_ablation} shows two ablations for parsing with gold EDUs. We see that both beam search and boundary LSTM (hierarchical encoding as shown in \Cref{fig:discourse_final_model_gold_edu_architecture}) are important to the model. The former can find better tree structure by searching a  larger searching space. The latter, meanwhile, connects the EDU-boundary representations, which enhances the model's ability to capture long-range dependencies between EDUs.

\begin{table}[t!]
\small 
\centering  
  \begin{tabular}{ l c c c c}
    \toprule
    Model  & Span & Nuc & Rel & Full \\
    \midrule
    Final model 			& 71.1  &59.6   &47.7   &46.8 \\
    \midrule
    \hspace{0.5em}\sout{Beam search}	& 70.1	&58.1	&46.8	&45.8 \\
    \hspace{0.5em}\sout{Boundary LSTM}	& 68.5	&55.5	&46.1	 &44.7 \\
    \bottomrule
  \end{tabular}
  \caption{\small Ablation test of our models \emph{with} gold EDUs. \sout{Beam search} indicates the full model with greedy decoding (beam width 1), while \sout{Boundary LSTM} is the full model with greedy decoding and no LSTM connection between EDU-boundary representations.}
  \label{table:gold_edu_ablation}
\normalsize  
\end{table}

\paragraph{End-to-end Parsing.}

For end-to-end parsing, \Cref{table:end2end_ablation} shows that the sentence boundary constraint (\Cref{sec:complete_models}) is indeed quite important to guide the model as it decodes long texts. Since sentence segmentation models are quite accurate, they can be employed if ground truth sentence segmentation is not available. We also notice that pretraining (GloVe) leads to improved performance.

\begin{table}[t!]
\small 
\centering  
\resizebox{\columnwidth}{!}{%
  \begin{tabular}{ l c c c c}
    \toprule
    Model  & Span & Nuc & Rel & Full \\
    \midrule
    Final model (GloVe) 			& 63.8  &53.0   &43.1   &42.1 \\
    \midrule
    \hspace{0.5em}\sout{GloVe}		& 63.3	&52.3	&42.4	&41.4 \\
    \hspace{0.5em}\sout{Sentence guidance}	& 59.2	&48.8	&40.7	 &38.9\\
    \hline
  \end{tabular}
  }
  \caption{\small Ablation test of our end-to-end model.
  \sout{GloVe} is the full model with randomized word embeddings, while
  \sout{Sentence guidance} is the full model with randomized word embeddings and without sentence guidance.
  }
  \label{table:end2end_ablation}
\normalsize  
\end{table}

\begin{figure}[t!]
\centering
\includegraphics[width=0.54\textwidth]{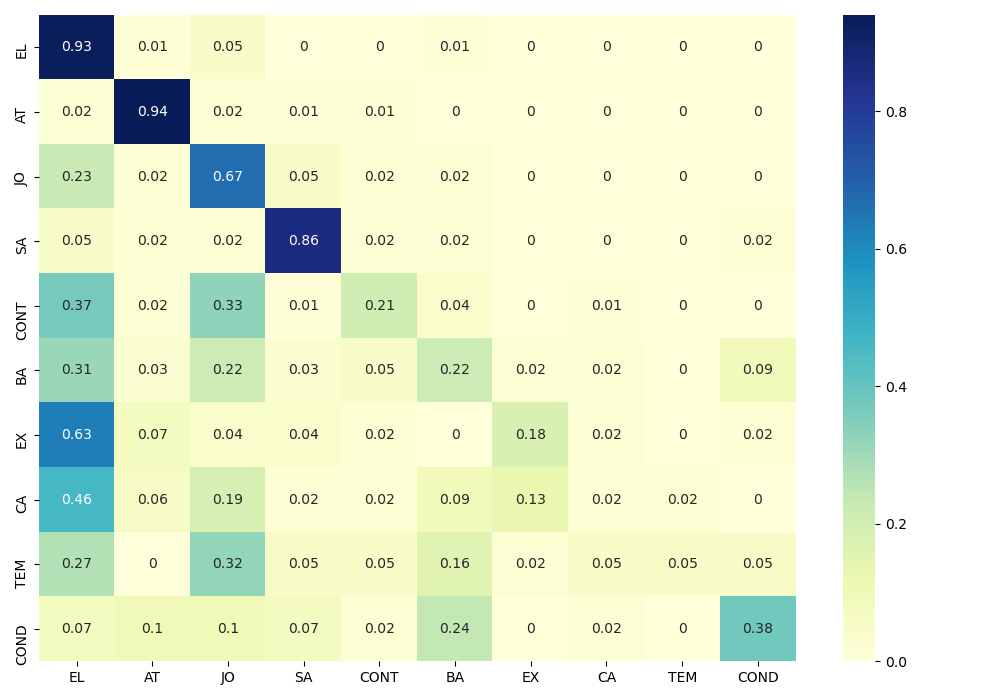}
\caption{\small Confusion matrix for the 10 most frequent relations on the RST–DT test set. The vertical axis represents true and horizontal axis represents predicted relations. The relations are: Elaboration (EL), Attribution (AT), Joint (JO), Same-Unit (SA), Contrast (CONT), Background (BA), Explanation (EX), Cause (CA), Temporal (TEM), Condition (COND).
}
\label{fig:confusion matrix}
\end{figure}

\begin{figure}[t!]
\centering
\includegraphics[width=0.5\textwidth]{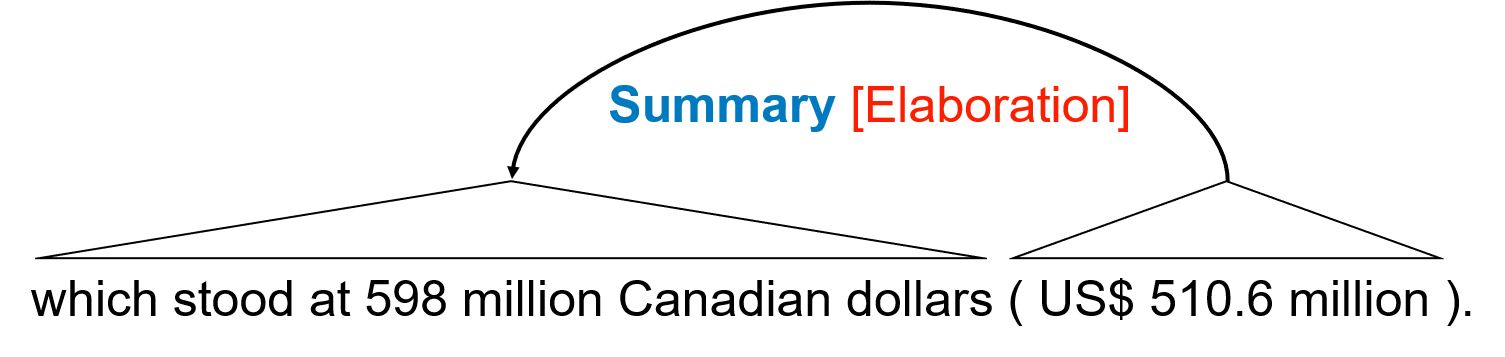}
\caption{\small An error example where our system incorrectly labels a Summary as Elaboration.
}
\label{fig:Mistake_Summary_as_Elaboration}
\end{figure}

\paragraph{Error Analysis.} We show our best parser's (with gold EDUs) confusion matrix for the 10 most frequent relation labels in \Cref{fig:confusion matrix}. The complete matrix with the 18 relations is shown in Appendix (\Cref{fig:full-confusion matrix}). The imbalanced relation distribution in RST-DT affects our model's performance to some extent. Also semantic similar relations tend to be confused with each other. \Cref{fig:Mistake_Summary_as_Elaboration} shows an example where our model mistakenly labels Summary as Elaboration. However, one could argue that the relation Elaboration is also valid here because the parenthesized text brings additional information (the equivalent amount of money). We show more error examples in the Appendix (\Cref{fig:Mistake_Condition_as_Background} - \ref{fig:Mistake_Explanation_as_Elaboration}), where our parser labels a Condition as Background, Temporal as Joint and Explanation as Elaboration. As we can see, all these relations are  semantically close and arguably interchangeable.

\begin{table}[tb]
\small 
\centering
\scalebox{0.75}{
\begin{tabular}{lcccc}  
\toprule
\textbf{System} & Gold Seg. & End-to-End&\bf{Time (s)} & \bf{Speedup}\\
\midrule
\citep{feng-hirst-2014-linear}  & \blackcheck& & 210 & 1.0x\\
\citep{yu-etal-2018-transition} & \blackcheck & &79 &2.7x\\
\midrule
Our parser (Glove) &\blackcheck & &19 & 11.1x\\
Our parser (XLNet) &\blackcheck & & 33 & 6.4x\\
Our parser (GloVe) & & \blackcheck& 45  & 4.7x\\
Our parser (XLNet) &  & \blackcheck &60 & 3.5x \\
\bottomrule
\end{tabular}
}
\caption{\small The wall time for parsing the RST-DT test set.
}
\label{table:speed}
\end{table}

\subsection{Parsing Speed}
Table \ref{table:speed} compares the parsing speed of our models with a representative non-neural \cite{feng-hirst-2014-linear} and neural model \cite{yu-etal-2018-transition}. We measure speed empirically using the wall time for parsing the test set. We ran the baselines and our models under the same settings (CPU: Intel Xeon W-2133 and GPU: Nvidia GTX 1080 Ti).

With gold-segmentation, our model with GloVe embeddings can parse the test set in 19 seconds, which is up to 11 times faster than \citep{feng-hirst-2014-linear}, and this is when their features are precomputed. The speed gain can be attributed to \Ni to the efficient GPU implementation of neural modules to process the decoding steps, and \Nii the fact that our model does not need to compute any handcrafted features. With pretrained models, our parser with gold segmentation is about 2.4 times faster than \citep{yu-etal-2018-transition}. Our end-to-end parser that also performs segmentation is faster than the baselines that are provided with the EDUs. 
Nonetheless, we believe there is still room for speed improvement by choosing a better network, like the Longformer \citep{Beltagy2020Longformer} which has an $\mathcal{O}(1)$ parallel time complexity in encoding a text, compared to the $\mathcal{O}(n)$ complexity of the recurrent encoder. 

\section{Related Work} \label{sec:rel}
Discourse analysis has been a long-established problem in NLP. Prior to the neural tsunami in NLP, discourse parsing methods commonly employed statistical models with handcrafted features \citep{soricut-marcu-2003-sentence,HernaultPdI10,feng-hirst-2014-linear,joty-etal-2015-codra}. Even within the neural paradigm, most previous studies still rely on external features to achieve their best performances \citep{ji-eisenstein-2014-representation,wang-etal-2017-two, braud-etal-2016-multi, braud-etal-2017-cross, yu-etal-2018-transition}. These parsers adopt a bottom-up approach, either transition-based or chart-based parsing.

Recently, top-down parsing has attracted more attention due to its ability to maintain an overall view of the input text. Inspired by the Stack-Pointer network \cite{ma-etal-2018-stack} for dependency parsing, \citet{lin-etal-2019-unified} first propose a seq2seq model for sentence-level parsing. \citet{zhang-etal-2020-top} extend this to the document level. \citet{Kobayashi2020TopDownRP} adopt a greedy splitting mechanism for discourse parsing inspired by \citet{minimal-span-based-parsing}'s work in constituency parsing. By using pretrained models/embeddings and extra features (\eg\ syntactic, text organizational features), these models achieve competitive results. However, their decoder infers a tree greedily.  

Our approach differs from previous work in that it can perform end-to-end discourse parsing in a single neural framework without needing segmentation as a prerequisite. Our model can parse a document from scratch without relying on any external features. Moreover, it can apply efficient beam search decoding to search for the best tree.

\section{Conclusion}

We have presented a novel top-down end-to-end method for discourse parsing based on a seq2seq model. Our model casts discourse parsing as a series of splitting decisions at token boundaries, which can solve discourse parsing and  segmentation in a single model. In both end-to-end parsing and parsing with gold segmentation, our parser achieves state-of-the-art, surpassing existing methods by a good margin, without relying on handcrafted features. Our parser is not only more effective but also more efficient than the existing ones.

This work leads us to several future directions. Our short-term goal is to improve the model with better architecture and training mechanisms. For example, joint training on discourse and syntactic parsing tasks could be a good future direction since both tasks are related and can be modeled within our unified conditional splitting framework. We also plan to extend our parser to other languages.  

\bibliographystyle{acl_natbib}
\bibliography{ref.bib}

\begin{thebibliography}{40}
\expandafter\ifx\csname natexlab\endcsname\relax\def\natexlab#1{#1}\fi

\bibitem[{Beltagy et~al.(2020)Beltagy, Peters, and
  Cohan}]{Beltagy2020Longformer}
Iz~Beltagy, Matthew~E. Peters, and Arman Cohan. 2020.
\newblock Longformer: The long-document transformer.
\newblock \emph{arXiv:2004.05150}.

\bibitem[{Bhatia et~al.(2015)Bhatia, Ji, and
  Eisenstein}]{bhatia-etal-2015-better}
Parminder Bhatia, Yangfeng Ji, and Jacob Eisenstein. 2015.
\newblock \href {https://doi.org/10.18653/v1/D15-1263} {Better document-level
  sentiment analysis from {RST} discourse parsing}.
\newblock In \emph{Proceedings of the 2015 Conference on Empirical Methods in
  Natural Language Processing}, pages 2212--2218, Lisbon, Portugal. Association
  for Computational Linguistics.

\bibitem[{Braud et~al.(2017)Braud, Coavoux, and
  S{\o}gaard}]{braud-etal-2017-cross}
Chlo{\'e} Braud, Maximin Coavoux, and Anders S{\o}gaard. 2017.
\newblock \href {https://www.aclweb.org/anthology/E17-1028} {Cross-lingual
  {RST} discourse parsing}.
\newblock In \emph{Proceedings of the 15th Conference of the {E}uropean Chapter
  of the Association for Computational Linguistics: Volume 1, Long Papers},
  pages 292--304, Valencia, Spain. Association for Computational Linguistics.

\bibitem[{Braud et~al.(2016)Braud, Plank, and
  S{\o}gaard}]{braud-etal-2016-multi}
Chlo{\'e} Braud, Barbara Plank, and Anders S{\o}gaard. 2016.
\newblock \href {https://www.aclweb.org/anthology/C16-1179} {Multi-view and
  multi-task training of {RST} discourse parsers}.
\newblock In \emph{Proceedings of {COLING} 2016, the 26th International
  Conference on Computational Linguistics: Technical Papers}, pages 1903--1913,
  Osaka, Japan. The COLING 2016 Organizing Committee.

\bibitem[{Carlson and Marcu(2001)}]{carlson-marcu-01}
Lynn Carlson and Daniel Marcu. 2001.
\newblock \href {http://www.isi.edu/~marcu/discourse/tagging-ref-manual.pdf}
  {Discourse tagging reference manual}.
\newblock Technical Report ISI-TR-545, University of Southern California
  Information Sciences Institute.

\bibitem[{Cross and Huang(2016)}]{cross-huang-2016-span}
James Cross and Liang Huang. 2016.
\newblock \href {https://doi.org/10.18653/v1/D16-1001} {Span-based constituency
  parsing with a structure-label system and provably optimal dynamic oracles}.
\newblock In \emph{Proceedings of the 2016 Conference on Empirical Methods in
  Natural Language Processing}, pages 1--11, Austin, Texas. Association for
  Computational Linguistics.

\bibitem[{Dozat and Manning(2017)}]{DozatM17}
Timothy Dozat and Christopher~D. Manning. 2017.
\newblock Deep biaffine attention for neural dependency parsing.
\newblock In \emph{5th International Conference on Learning Representations,
  {ICLR} 2017, Toulon, France, April 24-26, 2017, Conference Track
  Proceedings}.

\bibitem[{Feng and Hirst(2014)}]{feng-hirst-2014-linear}
Vanessa~Wei Feng and Graeme Hirst. 2014.
\newblock \href {https://doi.org/10.3115/v1/P14-1048} {A linear-time bottom-up
  discourse parser with constraints and post-editing}.
\newblock In \emph{Proceedings of the 52nd Annual Meeting of the Association
  for Computational Linguistics (Volume 1: Long Papers)}, pages 511--521,
  Baltimore, Maryland. Association for Computational Linguistics.

\bibitem[{Fisher and Roark(2007)}]{fisher-roark-2007-utility}
Seeger Fisher and Brian Roark. 2007.
\newblock \href {https://www.aclweb.org/anthology/P07-1062} {The utility of
  parse-derived features for automatic discourse segmentation}.
\newblock In \emph{Proceedings of the 45th Annual Meeting of the Association of
  Computational Linguistics}, pages 488--495, Prague, Czech Republic.
  Association for Computational Linguistics.

\bibitem[{Gao et~al.(2020)Gao, Wu, Li, Joty, Hoi, Xiong, King, and
  Lyu}]{gao-etal-2020-discern}
Yifan Gao, Chien-Sheng Wu, Jingjing Li, Shafiq Joty, Steven~C.H. Hoi, Caiming
  Xiong, Irwin King, and Michael Lyu. 2020.
\newblock \href {https://www.aclweb.org/anthology/2020.emnlp-main.191}
  {Discern: Discourse-aware entailment reasoning network for conversational
  machine reading}.
\newblock In \emph{Proceedings of the 2020 Conference on Empirical Methods in
  Natural Language Processing (EMNLP)}, pages 2439--2449, Online. Association
  for Computational Linguistics.

\bibitem[{Gerani et~al.(2014)Gerani, Mehdad, Carenini, Ng, and
  Nejat}]{gerani-etal-2014-abstractive}
Shima Gerani, Yashar Mehdad, Giuseppe Carenini, Raymond~T. Ng, and Bita Nejat.
  2014.
\newblock \href {https://doi.org/10.3115/v1/D14-1168} {Abstractive
  summarization of product reviews using discourse structure}.
\newblock In \emph{Proceedings of the 2014 Conference on Empirical Methods in
  Natural Language Processing ({EMNLP})}, pages 1602--1613, Doha, Qatar.
  Association for Computational Linguistics.

\bibitem[{Hernault et~al.(2010)Hernault, Prendinger, duVerle, and
  Ishizuka}]{HernaultPdI10}
Hugo Hernault, Helmut Prendinger, David~A. duVerle, and Mitsuru Ishizuka. 2010.
\newblock Hilda: A discourse parser using support vector machine
  classification.
\newblock \emph{Dialogue and Discourse}, 1(3):1--33.

\bibitem[{Hochreiter and Schmidhuber(1997)}]{Hochreiter:1997}
Sepp Hochreiter and J{\"u}rgen Schmidhuber. 1997.
\newblock Long short-term memory.
\newblock \emph{Neural computation}, 9(8):1735--1780.

\bibitem[{Ji and Eisenstein(2014)}]{ji-eisenstein-2014-representation}
Yangfeng Ji and Jacob Eisenstein. 2014.
\newblock \href {https://doi.org/10.3115/v1/P14-1002} {Representation learning
  for text-level discourse parsing}.
\newblock In \emph{Proceedings of the 52nd Annual Meeting of the Association
  for Computational Linguistics (Volume 1: Long Papers)}, pages 13--24,
  Baltimore, Maryland. Association for Computational Linguistics.

\bibitem[{Ji and Smith(2017)}]{ji-smith-2017-neural}
Yangfeng Ji and Noah~A. Smith. 2017.
\newblock \href {https://doi.org/10.18653/v1/P17-1092} {Neural discourse
  structure for text categorization}.
\newblock In \emph{Proceedings of the 55th Annual Meeting of the Association
  for Computational Linguistics (Volume 1: Long Papers)}, pages 996--1005,
  Vancouver, Canada. Association for Computational Linguistics.

\bibitem[{Joty et~al.(2012)Joty, Carenini, and Ng}]{joty-carenini-ng-emnlp-12}
Shafiq Joty, Giuseppe Carenini, and Raymond Ng. 2012.
\newblock \href {http://www.aclweb.org/anthology/D12-1083} {A novel
  discriminative framework for sentence-level discourse analysis}.
\newblock In \emph{Proceedings of the 2012 Joint Conference on Empirical
  Methods in Natural Language Processing and Computational Natural Language
  Learning}, EMNLP-CoNLL'12, pages 904--915, Jeju Island, Korea. ACL.

\bibitem[{Joty et~al.(2015)Joty, Carenini, and Ng}]{joty-etal-2015-codra}
Shafiq Joty, Giuseppe Carenini, and Raymond~T. Ng. 2015.
\newblock \href {https://doi.org/10.1162/COLI_a_00226} {{CODRA}: A novel
  discriminative framework for rhetorical analysis}.
\newblock \emph{Computational Linguistics}, 41(3):385--435.

\bibitem[{Joty et~al.(2013)Joty, Carenini, Ng, and
  Mehdad}]{joty-carenini-ng-mehdad-acl-13}
Shafiq Joty, Giuseppe Carenini, Raymond~T. Ng, and Yashar Mehdad. 2013.
\newblock {Combining Intra- and Multi-sentential Rhetorical Parsing for
  Document-level Discourse Analysis}.
\newblock In \emph{Proceedings of the 51st Annual Meeting of the Association
  for Computational Linguistics}, ACL'13, pages 486--496, Sofia, Bulgaria. ACL.

\bibitem[{Joty et~al.(2017)Joty, Guzm{\'a}n, M{\`a}rquez, and
  Nakov}]{joty-etal-2017-discourse}
Shafiq Joty, Francisco Guzm{\'a}n, Llu{\'\i}s M{\`a}rquez, and Preslav Nakov.
  2017.
\newblock \href {https://doi.org/10.1162/COLI_a_00298} {Discourse structure in
  machine translation evaluation}.
\newblock \emph{Computational Linguistics}, 43(4):683--722.

\bibitem[{Kingma and Ba(2015)}]{KingmaB14}
Diederik~P. Kingma and Jimmy Ba. 2015.
\newblock Adam: {A} method for stochastic optimization.
\newblock In \emph{3rd International Conference on Learning Representations,
  {ICLR} 2015, San Diego, CA, USA, May 7-9, 2015, Conference Track
  Proceedings}.

\bibitem[{Kobayashi et~al.(2020)Kobayashi, Hirao, Kamigaito, Okumura, and
  Nagata}]{Kobayashi2020TopDownRP}
Naoki Kobayashi, Tsutomu Hirao, Hidetaka Kamigaito, Manabu Okumura, and Masaaki
  Nagata. 2020.
\newblock Top-down rst parsing utilizing granularity levels in documents.
\newblock In \emph{Proceedings of the 2020 Conference on Artificial
  Intelligence for the American (AAAI)}, pages 8099--8106.

\bibitem[{Li et~al.(2018)Li, Sun, and Joty}]{ijcai2018-579}
Jing Li, Aixin Sun, and Shafiq Joty. 2018.
\newblock \href {https://doi.org/10.24963/ijcai.2018/579} {Segbot: A generic
  neural text segmentation model with pointer network}.
\newblock In \emph{Proceedings of the Twenty-Seventh International Joint
  Conference on Artificial Intelligence, {IJCAI-18}}, pages 4166--4172.
  International Joint Conferences on Artificial Intelligence Organization.

\bibitem[{Li et~al.(2016)Li, Li, and Chang}]{li-etal-2016-discourse}
Qi~Li, Tianshi Li, and Baobao Chang. 2016.
\newblock \href {https://doi.org/10.18653/v1/D16-1035} {Discourse parsing with
  attention-based hierarchical neural networks}.
\newblock In \emph{Proceedings of the 2016 Conference on Empirical Methods in
  Natural Language Processing}, pages 362--371, Austin, Texas. Association for
  Computational Linguistics.

\bibitem[{Lin et~al.(2019)Lin, Joty, Jwalapuram, and
  Bari}]{lin-etal-2019-unified}
Xiang Lin, Shafiq Joty, Prathyusha Jwalapuram, and M~Saiful Bari. 2019.
\newblock \href {https://doi.org/10.18653/v1/P19-1410} {A unified linear-time
  framework for sentence-level discourse parsing}.
\newblock In \emph{Proceedings of the 57th Annual Meeting of the Association
  for Computational Linguistics}, pages 4190--4200, Florence, Italy.
  Association for Computational Linguistics.

\bibitem[{Lynn et~al.(2002)Lynn, Marcu, and Okurowski.}]{RST:DT:2002}
Carlson Lynn, Daniel Marcu, and Mary~Ellen Okurowski. 2002.
\newblock Rst discourse treebank (rst--dt) ldc2002t07.
\newblock \emph{Linguistic Data Consortium}.

\bibitem[{Ma et~al.(2018)Ma, Hu, Liu, Peng, Neubig, and
  Hovy}]{ma-etal-2018-stack}
Xuezhe Ma, Zecong Hu, Jingzhou Liu, Nanyun Peng, Graham Neubig, and Eduard
  Hovy. 2018.
\newblock \href {https://doi.org/10.18653/v1/P18-1130} {Stack-pointer networks
  for dependency parsing}.
\newblock In \emph{Proceedings of the 56th Annual Meeting of the Association
  for Computational Linguistics (Volume 1: Long Papers)}, pages 1403--1414,
  Melbourne, Australia. Association for Computational Linguistics.

\bibitem[{Maas et~al.(2013)Maas, Hannun, and
  Ng}]{Maas13rectifiernonlinearities}
Andrew~L. Maas, Awni~Y. Hannun, and Andrew~Y. Ng. 2013.
\newblock Rectifier nonlinearities improve neural network acoustic models.
\newblock In \emph{in ICML Workshop on Deep Learning for Audio, Speech and
  Language Processing}.

\bibitem[{Mann and Thompson(1988)}]{mann1988rhetorical}
William~C Mann and Sandra~A Thompson. 1988.
\newblock Rhetorical structure theory: Toward a functional theory of text
  organization.
\newblock \emph{Text}, 8(3):243--281.

\bibitem[{Marcu(2000)}]{marcu-2000-rhetorical}
Daniel Marcu. 2000.
\newblock \href {https://www.aclweb.org/anthology/J00-3005} {The rhetorical
  parsing of unrestricted texts: a surface-based approach}.
\newblock \emph{Computational Linguistics}, 26(3):395--448.

\bibitem[{Morey et~al.(2017)Morey, Muller, and Asher}]{morey-etal-2017-much}
Mathieu Morey, Philippe Muller, and Nicholas Asher. 2017.
\newblock \href {https://doi.org/10.18653/v1/D17-1136} {How much progress have
  we made on {RST} discourse parsing? a replication study of recent results on
  the {RST}-{DT}}.
\newblock In \emph{Proceedings of the 2017 Conference on Empirical Methods in
  Natural Language Processing}, pages 1319--1324, Copenhagen, Denmark.
  Association for Computational Linguistics.

\bibitem[{Morey et~al.(2018)Morey, Muller, and
  Asher}]{morey-etal-2018-dependency}
Mathieu Morey, Philippe Muller, and Nicholas Asher. 2018.
\newblock \href {https://doi.org/10.1162/COLI_a_00314} {A dependency
  perspective on {RST} discourse parsing and evaluation}.
\newblock \emph{Computational Linguistics}, 44(2):197--235.

\bibitem[{Pennington et~al.(2014)Pennington, Socher, and
  Manning}]{pennington2014glove}
Jeffrey Pennington, Richard Socher, and Christopher~D. Manning. 2014.
\newblock \href {http://www.aclweb.org/anthology/D14-1162} {Glove: Global
  vectors for word representation}.
\newblock In \emph{Empirical Methods in Natural Language Processing (EMNLP)},
  pages 1532--1543.

\bibitem[{Soricut and Marcu(2003)}]{soricut-marcu-2003-sentence}
Radu Soricut and Daniel Marcu. 2003.
\newblock \href {https://www.aclweb.org/anthology/N03-1030} {Sentence level
  discourse parsing using syntactic and lexical information}.
\newblock In \emph{Proceedings of the 2003 Human Language Technology Conference
  of the North {A}merican Chapter of the Association for Computational
  Linguistics}, pages 228--235.

\bibitem[{Stern et~al.(2017)Stern, Andreas, and
  Klein}]{minimal-span-based-parsing}
Mitchell Stern, Jacob Andreas, and Dan Klein. 2017.
\newblock A minimal span-based neural constituency parser.
\newblock In \emph{Proceedings of the 55th Annual Meeting of the Association
  for Computational Linguistics, {ACL} 2017, Vancouver, Canada, July 30 -
  August 4, Volume 1: Long Papers}, pages 818--827.

\bibitem[{Vinyals et~al.(2015)Vinyals, Fortunato, and Jaitly}]{VinyalsNIPS2015}
Oriol Vinyals, Meire Fortunato, and Navdeep Jaitly. 2015.
\newblock \href {http://papers.nips.cc/paper/5866-pointer-networks.pdf}
  {Pointer networks}.
\newblock In C.~Cortes, N.~D. Lawrence, D.~D. Lee, M.~Sugiyama, and R.~Garnett,
  editors, \emph{Advances in Neural Information Processing Systems 28}, pages
  2692--2700. Curran Associates, Inc.

\bibitem[{Wang et~al.(2017)Wang, Li, and Wang}]{wang-etal-2017-two}
Yizhong Wang, Sujian Li, and Houfeng Wang. 2017.
\newblock \href {https://doi.org/10.18653/v1/P17-2029} {A two-stage parsing
  method for text-level discourse analysis}.
\newblock In \emph{Proceedings of the 55th Annual Meeting of the Association
  for Computational Linguistics (Volume 2: Short Papers)}, pages 184--188,
  Vancouver, Canada. Association for Computational Linguistics.

\bibitem[{Yang et~al.(2019)Yang, Dai, Yang, Carbonell, Salakhutdinov, and
  Le}]{NEURIPS2019_dc6a7e65}
Zhilin Yang, Zihang Dai, Yiming Yang, Jaime Carbonell, Russ~R Salakhutdinov,
  and Quoc~V Le. 2019.
\newblock \href
  {https://proceedings.neurips.cc/paper/2019/file/dc6a7e655d7e5840e66733e9ee67cc69-Paper.pdf}
  {Xlnet: Generalized autoregressive pretraining for language understanding}.
\newblock In \emph{Advances in Neural Information Processing Systems},
  volume~32, pages 5753--5763. Curran Associates, Inc.

\bibitem[{Yu et~al.(2018)Yu, Zhang, and Fu}]{yu-etal-2018-transition}
Nan Yu, Meishan Zhang, and Guohong Fu. 2018.
\newblock \href {https://www.aclweb.org/anthology/C18-1047} {Transition-based
  neural {RST} parsing with implicit syntax features}.
\newblock In \emph{Proceedings of the 27th International Conference on
  Computational Linguistics}, pages 559--570, Santa Fe, New Mexico, USA.
  Association for Computational Linguistics.

\bibitem[{Zeldes et~al.(2019)Zeldes, Das, Maziero, Antonio, and
  Iruskieta}]{zeldes-etal-2019-disrpt}
Amir Zeldes, Debopam Das, Erick~Galani Maziero, Juliano Antonio, and Mikel
  Iruskieta. 2019.
\newblock \href {https://doi.org/10.18653/v1/W19-2713} {The {DISRPT} 2019
  shared task on elementary discourse unit segmentation and connective
  detection}.
\newblock In \emph{Proceedings of the Workshop on Discourse Relation Parsing
  and Treebanking 2019}, pages 97--104, Minneapolis, MN. Association for
  Computational Linguistics.

\bibitem[{Zhang et~al.(2020)Zhang, Xing, Kong, Li, and
  Zhou}]{zhang-etal-2020-top}
Longyin Zhang, Yuqing Xing, Fang Kong, Peifeng Li, and Guodong Zhou. 2020.
\newblock \href {https://doi.org/10.18653/v1/2020.acl-main.569} {A top-down
  neural architecture towards text-level parsing of discourse rhetorical
  structure}.
\newblock In \emph{Proceedings of the 58th Annual Meeting of the Association
  for Computational Linguistics}, pages 6386--6395, Online. Association for
  Computational Linguistics.

\end{thebibliography}
\section*{Appendix}
\section{Parsing with EDUs}
\begin{figure}[h!]
\centering
\includegraphics[width=0.8\textwidth]{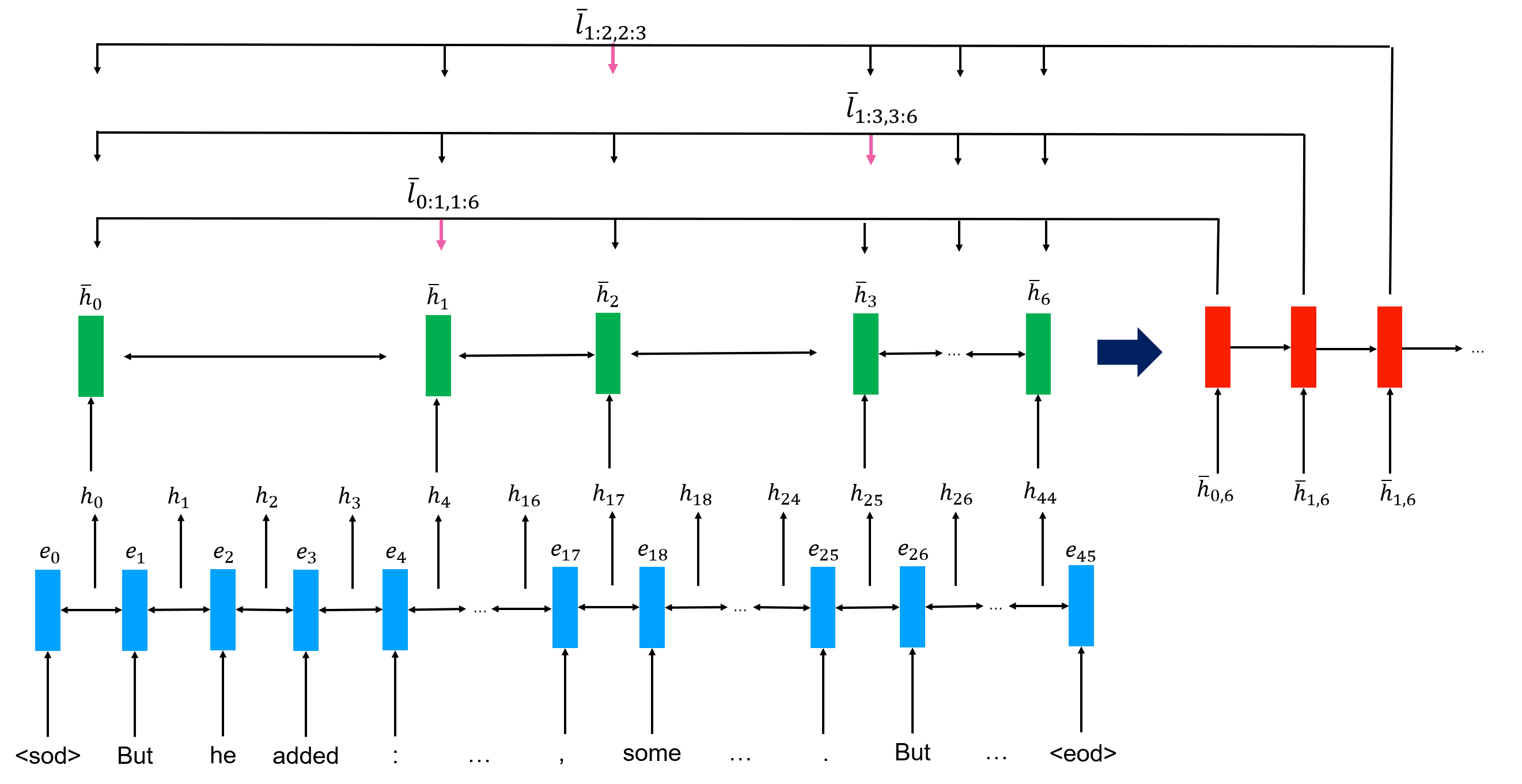}
\caption{Our discourse parser along with the first few decoding steps. When EDUs are given, we use a hierarchical EDU encoding (Boundary LSTM) to model EDU boundary representations.}
\label{fig:discourse_final_model_gold_edu_architecture}
\end{figure}

Figure \ref{fig:discourse_final_model_gold_edu_architecture} shows first few decoding steps with final parsers with EDUs.\\

\section{Error Analysis} 
\begin{figure*}[t!]
\centering
\includegraphics[width=0.9\textwidth]{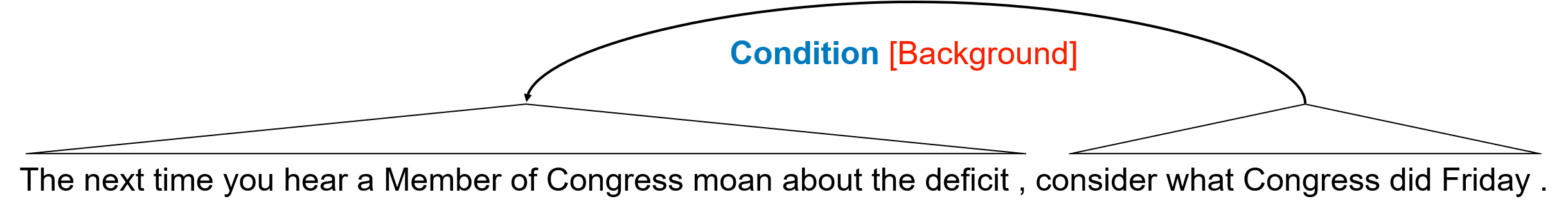}
\caption{\small Our system incorrectly labels a Condition as Background.
}
\label{fig:Mistake_Condition_as_Background}
\end{figure*}

\begin{figure*}[t!]
\centering
\includegraphics[width=0.9\textwidth]{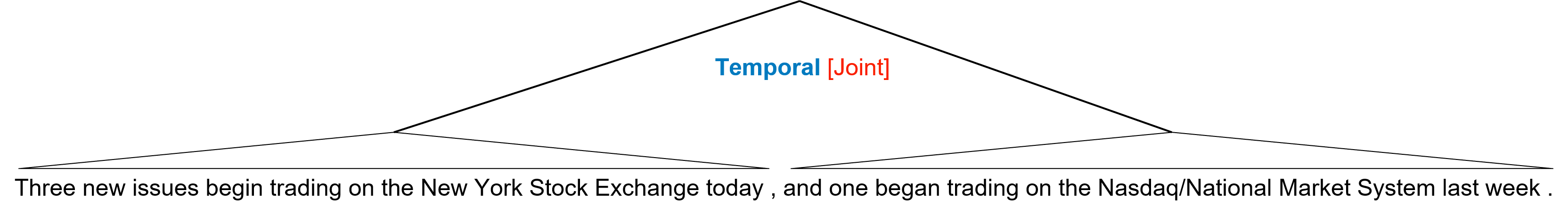}
\caption{\small Our system incorrectly labels a Temporal as Joint.
}
\label{fig:Mistake_Mistake_Temporal_as_Joint}
\end{figure*}

\begin{figure*}[t!]
\centering
\includegraphics[width=0.9\textwidth]{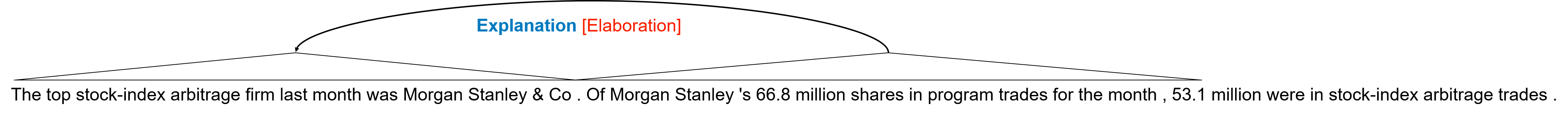}
\caption{\small Our system incorrectly labels a Explanation as Elaboration.
}
\label{fig:Mistake_Explanation_as_Elaboration}
\end{figure*}

We show our best parser's (with gold EDUs) confusion matrix for all relation labels in \Cref{fig:confusion matrix}.  \Cref{fig:Mistake_Condition_as_Background} -  \ref{fig:Mistake_Explanation_as_Elaboration} present examples where our  parser falsely labels a Condition as Background, Temporal as Joint and Explanation as Elaboration.

\begin{figure*}[t!]
\centering
\includegraphics[width=0.9\textwidth]{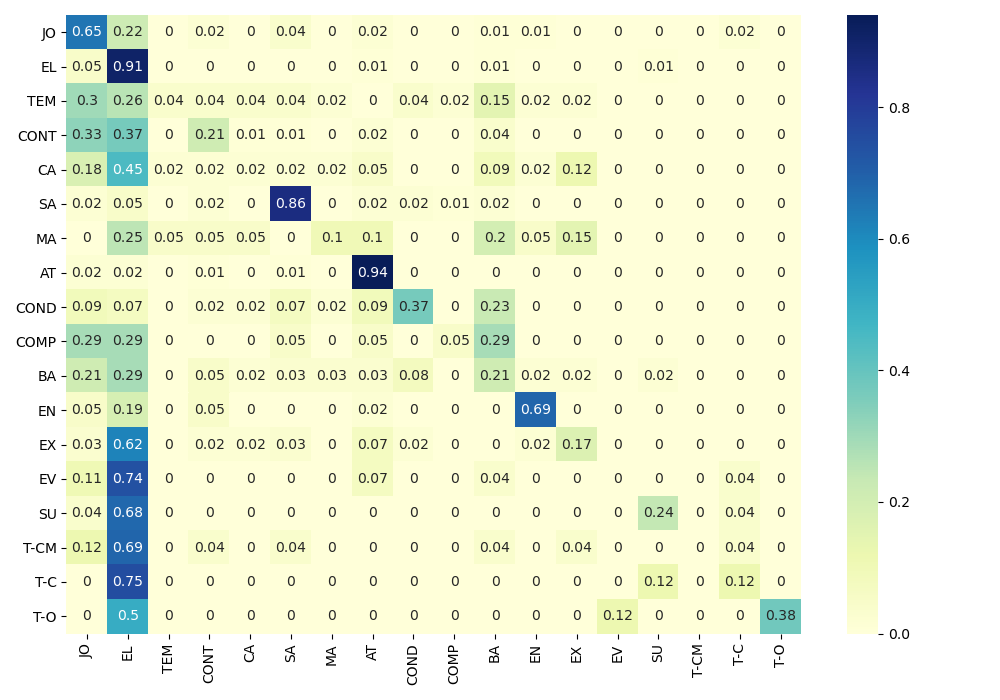}
\caption{\small Confusion matrix for  relation labels on the RST–DT test set. The vertical axis represents true and horizontal axis represents predicted relations. The relations are: Joint (JO), Elaboration (EL), Temporal (TEM), Contrast (CONT), Cause (CA), Same-Unit (SA), Manner-Means (MA), Attribution (AT), 
Condition (COND), Comparison (COMP), Background (BA), Enablement (EN), Explanation (EX), Evaluation (EV),
Summary (SU), Topic-Comment (T-CM), Topic-Change (T-C) and TextualOrganization (T-O).
}
\label{fig:full-confusion matrix}
\end{figure*}
\vspace{-0.5em}

\vspace{-0.5em}
\vspace{-0.5em}
\vspace{-0.5em}
\end{document}